\documentclass{llncs}

\usepackage{hyperref}

\usepackage[utf8]{inputenc}
\usepackage[english]{babel}
\usepackage{amssymb}
\usepackage{amsmath}
\usepackage{latexsym}
\usepackage{amsfonts}
\usepackage{wrapfig} 
\usepackage{graphicx}
\usepackage{nccmath}
\usepackage{blkarray}
\usepackage{multirow}
\usepackage{times}

\let\llncssubparagraph\subparagraph
\let\subparagraph\paragraph
\usepackage[compact]{titlesec}
\let\subparagraph\llncssubparagraph

 \usepackage[compact]{titlesec}
    \titlespacing{\section}{0pt}{3ex}{1ex}
    \titlespacing{\subsection}{0pt}{2ex}{1ex}
    \titlespacing{\subsubsection}{0pt}{0.5ex}{0ex}
    
\usepackage{setspace}
    {\end{pmatrix}\end{medsize}}%

\usepackage{tikz}
\usepackage{cite}
\usetikzlibrary{arrows,shapes,automata,backgrounds,petri,matrix,calc,patterns,fit}

\DeclareGraphicsExtensions{.pdf,.jpeg,.png}
\usepackage{graphicx}
\usepackage{caption}
\usepackage{subcaption}
\usepackage{enumerate}
\usepackage{float}
\usepackage{color,soul}
\usepackage{array,colortbl,xcolor}
\usetikzlibrary{tikzmark}
\usetikzlibrary{decorations.pathmorphing,shapes}

\usepackage[framemethod=tikz]{mdframed}
\usepackage{pdflscape}
\usepackage{todonotes}
\usepackage{algorithm}
\usepackage{algpseudocode}
\usepackage{adjustbox}
\usepackage{wrapfig,graphicx,lipsum}

\newcounter{sarrow}

\title{Business Process Variant Analysis based on\\Mutual Fingerprints of Event Logs}
\author{Farbod Taymouri$^{1}$, Marcello La Rosa$^1$, Josep Carmona$^2$}
\institute{
$^1$The University of Melbourne, Melbourne (Australia)\\
$^2$ Universitat Polit\`ecnica de Catalunya, Barcelona (Spain)
\\
\email{\{farbod.taymouri, marcello.larosa\}@unimelb.edu.au \\
jcarmona@cs.upc.edu} 
}
\begin{document}
\maketitle
	
\begin{abstract}Comparing business process variants using event logs is a common use case in process mining. Existing techniques for process variant analysis detect statistically-significant differences between variants at the level of individual entities (such as process activities) and their relationships (e.g.\  directly-follows relations between activities). This may lead to a proliferation of differences due to the low level of granularity in which such differences are captured. This paper presents a novel approach to detect statistically-significant differences between variants at the level of entire process traces (i.e.\ sequences of directly-follows relations). 
The cornerstone of this approach is a technique to learn a directly-follows graph called {\em mutual fingerprint} from the event logs of the two variants. A mutual fingerprint is a lossless encoding of a set of traces and their duration using discrete wavelet transformation. This structure facilitates the understanding of statistical differences along the control-flow and performance dimensions.
The approach has been evaluated using real-life event logs against two baselines. The results show that at a trace level, the baselines cannot always reveal the differences discovered by our approach, or can detect spurious differences. 
\end{abstract}


\section{Introduction}
\label{sec:introduction}



The complexity of modern organizations leads to the co-existence of different \textit{variants} of the same business process. Process variants may be determined based on different logical drivers, such as brand, product, type of customer, geographic location, as well as performance drivers, e.g.\ cases that complete on-time vs. cases that are slow.

Identifying and explaining differences between process variants can help not only in the context of process standardization initiatives, but also to identify root causes for performance deviations or compliance violations. 
For example, in the healthcare domain, two patients with similar diseases might experience different care pathways, even if they are supposed to be treated alike \cite{Suriadi2014,Swinnen2012}. Moreover, even if the care pathways are the same in terms of sequences of activities, they could have different performance, e.g. one patient may be discharged in a much shorter timeframe than the other \cite{Poelmans2010}. 


\begin{figure}[h]
\vspace{-5mm}
	\centering
	\includegraphics[width=1\linewidth]{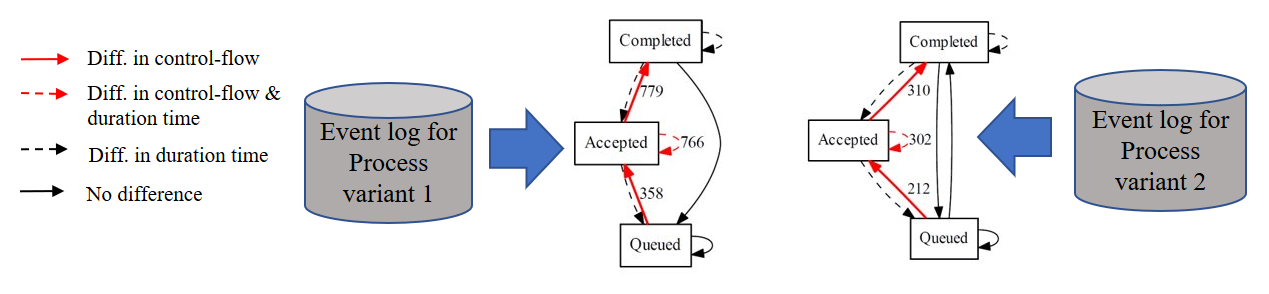}
	\vspace{-2mm}
	\caption{\small{Example of mutual fingerprints for an IT service desk process}}
	\label{fig:finger example}
\vspace{-8mm}
\end{figure}

Comparing business process variants using process execution data (a.k.a. event logs) recorded by information systems, is a common use case in process mining \cite{Aalst2016}. Existing techniques for process variant analysis \cite{bolt2018, Nguyen2018-Multiperspectquteprints117962} detect statistically-significant differences between variants at the level of individual entities (such as process activities) and their relationships (e.g.\ directly-follows relations between activities). However, these techniques often lead to a proliferation of differences due to the low level of granularity in which such differences are captured.

This paper presents a statistically sound approach for process variant analysis, that examines both the order in which process activities are executed (a.k.a. control-flow) and their duration (a.k.a. performance), using the event logs of two process variants as input. The cornerstone feature of this approach is the ability to provide statistically significant control-flow differences between process variants, 
via the use of a novel graph-based representation of a process, namely {\em mutual fingerprint}. The fingerprint of a process variant is a directly-follows graph that only shows the behavior of that variant that is statistically significantly different 
from that of another process variant, hence the term ``mutual''. This lossless encoding of differences can be seen as the latent representation of a process variant that provides a considerably simplified representation of the underlying behavior, focusing only on differences. For example, Fig. \ref{fig:finger example} shows the discovered mutual fingerprints for two example process variants. One can see that the fingerprint of process variant 2 has an extra edge, i.e., \emph{(Queued, Completed)}, that 
does not appear in the other fingerprint. In a mutual fingerprint graph, different edge types are used to capture differences in the control-flow and activity duration.

The approach to construct mutual fingerprints consists of three steps: i) feature generation, ii) feature selection, and iii) filtering. Given the event log of the two variants, the first step exploits {\em Discrete Wavelet Transformation} to obtain a lossless encoding of the two variants 
(along the control-flow and activity duration dimensions) into a set of vectors. The second step
adopts a machine learning strategy combined with statistical tests to determine what subset of features, i.e.\ events, discriminates the two process variants at a given significant level. The third step filters traces of each process variant that do not carry any discriminatory events. 


The approach has been evaluated using process variants from four real-life event logs, against two baseline techniques. The comparison includes a quantitative assessment of the results and a discussion on execution time.  

The paper is organized as follows. Related work and preliminaries are presented in Section \ref{sec:related work}, and \ref{sec:preliminaries} respectively. The approach is presented in Section \ref{sec: proposed approach}, followed by the evaluation in Section \ref{sec:Experiment}. Finally, Section \ref{sec:conclusion} concludes the paper and discusses some avenues for future work.

\section{Related Work}
\label{sec:related work}
We report only the most recent approaches related to our contribution. The interested reader can find a complete analysis in~\cite{taymouri2019business}. 
The work by van Beest et al. \cite{Beest2015} relies on the product automaton of two event structures to distill all the behavioral differences between two process variants from the respective event logs, and render these differences to end users via textual explanations.  
Cordes et al. \cite{Cordes2015} discover two process models and their differences are defined as the minimum number of operations that transform on model to the other. This work was extended in \cite{Ballambettu2017} to compare process variants using annotated transition systems. Similarly, \cite{Kriglstein13} creates a difference model between two input process models to represent differences.
Pini et al. \cite{Pini2015} contribute a visualization technique that compares two discovered process models in terms of performance data. The work in \cite{WYNN201793} proposes an extension of this work, by considering a normative process model alongside with event logs as inputs, and adding more data preparation facilities. Similarly, \cite{LOW2017106}   
 develops visualisation techniques to provide targeted analysis of resource reallocation and activity rescheduling.

Particularly relevant to our approach are the works by Bolt et al. and Nguyen et al., because they are grounded on statistical significance. Bolt et al. \cite{Bolt2016} use an annotated transition system to highlight the differences between process variants. The highlighted parts only show different dominant behaviors that are statistically significant with respect to edge frequencies. 
This work was later extended 
in \cite{bolt2018}, by inducting decision trees for performance data among process variants. Nguyen et al. \cite{Nguyen2018-Multiperspectquteprints117962} encode process variants into  {\em Perspective Graphs}. The comparison of perspective graphs results in a {\em Differential Graph}, which is a graph that contains common nodes and edges, and also nodes and edges that appear in one perspective graph only. As shown in the evaluation carried out in this paper, these two works, while relying on statistical tests, may lead to a proliferation of differences due to the low level in which such differences are captured (individual activities or activity relations). Our approach lifts these limitations by extracting entire control-flow paths or performance differences that constitute statistically significant differences between the two process variants. 

\section{Preliminaries}
\label{sec:preliminaries}

In this section we introduce preliminary definitions required to describe our approach such as event, trace, event log and process variant. Next, we provide some basic linear algebra definitions that will be specifically used for our featuring encoding.




\begin{definition}[Event, Trace, Event Log] An $event$ is a tuple $(a, c, t, (d_1, v_1), \ldots, (d_m, \\v_m))$ where $a$ is the activity name, $c$ is the case id, $t$ is the timestamp and $(d_1, v_1), \ldots, \\(d_m, v_m)$ (where $m \geq 0$) are the event or case attributes and their values.
A $trace$ is a non-empty sequence $\sigma = e_1,\ldots,e_{n}$ of events such that $\forall i,j \in [1..n] \; e_i{.}c = e_j{.}c$. An event log $L$ is a set  $\sigma_1, \ldots \sigma_n$ of traces.

\end{definition}

\begin{definition}[Process variant]
An event log $L$ can be partitioned into a finite set of groups called  process variants $\varsigma_1, \varsigma_2, \dots, \varsigma_n$, such that 
$\exists d$ such that $\forall$ $\varsigma_k$ and  $\forall \sigma_i, \sigma_j \in \varsigma_k$, $\sigma_i.d = \sigma_j.d$.
\label{def:variant}
\end{definition}

\noindent The above definition of a process variant emphasizes that process executions in the same group must have the same  value for a given attribute, and each process execution belongs only to one process variant\footnote{Definition~\ref{def:variant} can be easily generalized to more than one attribute, and arbitrary comparisons.}.

\begin{definition}[Vector]
\label{def:vector}
A vector, $\mathbf{x} = (x_1,x_2,\dots, x_n)^T$, is a column array of elements where 
the $ith$ element is shown by $x_i$. 
If each element is in $\mathbb{R}$ and vector contains $n$ elements, then the vector lies in $\mathbb{R}^{n \times 1}$, and the dimension of $\mathbf{x}$, $dim(\mathbf{x})$, is $n\times 1$.
\end{definition}

\noindent We represent a set of $d$ vectors as $\mathbf{x}^{(1)}, \mathbf{x}^{(2)}, \dots, \mathbf{x}^{(d)}$, where $x^{(i)} \in \mathbb{R}^{n \times 1}$. Also, they can be represented by a matrix $\mathbf{M} = ( \mathbf{x}^{(1)}, \mathbf{x}^{(2)}, \dots, \mathbf{x}^{(d)} )$ where $\mathbf{M} \in \mathbb{R}^{n\times d}$. We denote the $ith$ row of a matrix by $\mathbf{M}_{i,:}$, and likewise the $ith$ column by $\mathbf{M}_{:,i}$. The previous definitions can be extended for a set of columns or rows, for example if $R=\{3,5,9\}$ and $C=\{1,4,6,12\}$, then $\mathbf{M}_{R,C}$ returns the indicated rows and columns.


\begin{definition}[Vector space]
\label{def:vector space}
A vector space consists of a set $V$ of vectors, a field $\mathbb{F}$ ($\mathbb{R}$ for real numbers), and two operations $+,\times$ with the following properties, $\forall \mathbf{u,v} \in V, \mathbf{u}+\mathbf{v} \in V$, and $\forall c\in F, \forall \mathbf{v} \in V, c \times \mathbf{v} \in V$.
\end{definition}


\begin{definition}[Basis vectors]
\label{def:basis vectors}
A set $B$ of vectors in a vector space $V$ is called a basis, if every element of $V$ can be written as a finite linear combination of elements of $B$. The coefficients of this linear combination are referred to as coordinates on $B$ of the vector.
\end{definition}

A set $B$ of basis vectors is called \emph{orthonormal}, if $\forall \mathbf{u,v} \in B, <\mathbf{u}^T,\mathbf{v}>=0 $, and $\|\mathbf{u}\|_2=1$, $\|\mathbf{v}\|_2=1$. A \emph{basis matrix} is a matrix whose columns are basis vectors.


For example, the set of $\mathbf{e}^{(1)} = (0,0,0,1)^T$, $\mathbf{e}^{(2)} = (0,0,1,0)^T$, $\mathbf{e}^{(3)} = (0,1,0,0)^T$, and $\mathbf{e}^{(4)} = (1,0,0,0)^T$ constitutes a vector space in $\mathbb{R}^4$. Also, they are orthonormal basis vectors in $\mathbb{R}^4$, since every vector in that space can be represented by finite combination of them, for example, $(1,2,-3,4)^T = 1 \times \mathbf{e}^{(4)}+ 2 \times \mathbf{e}^{(3)} - 3 \times \mathbf{e}^{(2)} + 4 \times \mathbf{e}^{(1)}$. The set of  $\mathbf{e}^{(1)},\mathbf{e}^{(2)},\mathbf{e}^{(3)}$, and $\mathbf{e}^{(4)}$ are linearly independent. The corresponding basis matrix, called \emph{canonical basis} matrix, is $\mathbf{E} = (\mathbf{e}^{(1)}, \mathbf{e}^{(2)}, \mathbf{e}^{(3)},\mathbf{e}^{(4)})$.

\vspace{-2mm}
\section{Proposed Approach}
\label{sec: proposed approach}

Process variant analysis can help business analysts to find \textit{why} and \textit{how} two business process variants, each represented by a set of process executions, differ from each other. In this paper we focus on statistically identifying the differences of two process variants, either in the control flow or in the performance dimension. For instance, we are interested in identifying which sequences of activities occur more frequently in one process variant, or which activity has a statistically significant difference in duration between the two process variants.


\begin{figure}[h]
\vspace{-5mm}
	\centering
	\includegraphics[width=.8\linewidth]{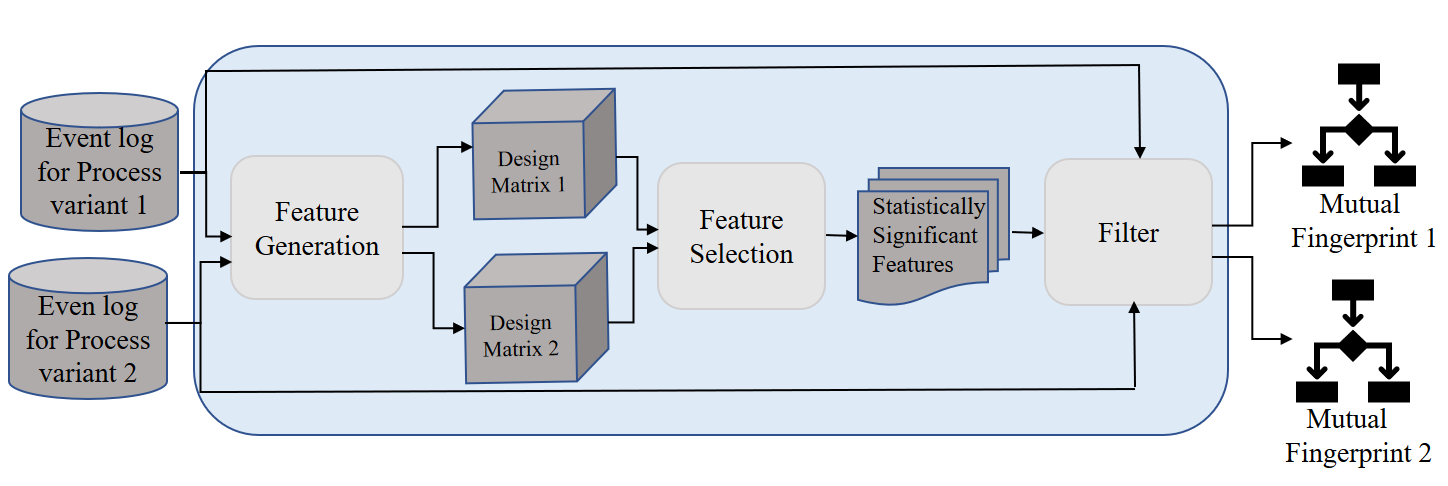}
	\caption{Approach for constructing mutual fingerprints}
	\label{fig:framework}
	\vspace{-5mm}
\end{figure}

Given the event logs of two process variants, our approach revolves around the construction of a representative directly-follows graph from each variant, called \emph{mutual fingerprint}. A fingerprint highlights the statistically-significant differences between the two variants, along the control-flow and activity duration dimensions. To construct such mutual fingerprints, we perform the following three steps, as shown in Fig. \ref{fig:framework}:



\begin{enumerate}
    \item \textbf{Feature generation}: This step encodes every single trace of the input event log of each of the two process variants, into a set of vectors of the same length for every event. Each vector contains the respective wavelet coefficients for a specific event. Essentially, a wavelet coefficient is an encoding of the time series behind each trace. For a trace, these vectors are stacked into a single vector. This way allows the encoding of a process variant as a matrix, called \textit{design matrix}, which is used in the next step.

    
    \item \textbf{Feature selection}: In this step, the wavelet coefficients are used to build features to train a binary classifier. 
    This procedure can be repeated several times for cross-validation purposes, and a statistical test is performed on top of the results of cross-validation, to ensure that the selected features (events classes in the log) provide enough information to discriminate the two classes arising from the two process variants.
    

    \item \textbf{Filtering}: This last step filters the log of each process variant by keeping only those traces that contain discriminatory events discovered from the previous stage. A mutual fingerprint is then built from the traces left for each process variant log.
    
\end{enumerate}

In the rest of this section, we first formally define the notion of discrete wavelet transformation and then use this to illustrate the above three steps in detail.

\subsection{Discrete Wavelet Transformation and Time-Series Encoding}

In Sec. \ref{sec:preliminaries} we defined a vector space, however, it must be noted that for an arbitrary vector space, there are infinitely number of basis matrices where one can be obtained from the others by a linear transformation \cite{strang16}. Among several basis matrices, \emph{Haar basis}  matrix is one of the most important set of basis matrix in $\mathbb{R}^n$ that plays an important role in analysing sequential data \cite{HaarWaveletApplication}. Formally, it is defined as follows (for the sake of exposition lets assume that the dimension is power of two):

\begin{definition}[Haar basis matrix]
\label{def:haar basis vectors}
Given a dimension of power two, i.e., $2^n$, the Haar basis matrix can be represented by the following recurrent equation \cite{WaveletRecurrent}:
\small{
\begin{equation}
 \mathbf{H}(n) = \begin{pmatrix}
       \mathbf{H}(n-1) & \otimes \begin{pmatrix}
1\\
1
\end{pmatrix} ,    \mathbf{I}(n-1) & \otimes \begin{pmatrix}
1\\
-1
\end{pmatrix}      \\[0.3em]
     \end{pmatrix}, \quad \mathbf{H}(0) =1
\end{equation}
}

\normalsize

\noindent where $\mathbf{H}(n)$ is the matrix of Haar vectors of degree $2^n$, $\mathbf{I}(n)$ is an identity matrix of size $2^n$, and $\otimes$ is the outer-product operator.
\end{definition}

 Haar basis vectors can be derived for dimensions of arbitrary lengths that are not necessarily to be power of two, however, the recurrent formula becomes more complicated \cite{GallieLinearAlgebra2019book}. A few examples of Haar basis matrices are as follows:
 
 \small{
\begin{equation*}
    \mathbf{H}(1) = \begin{pmatrix}
       1 & 1           \\[0.3em]
       1 & -1
     \end{pmatrix}, \quad     \mathbf{H}(2) = \begin{pmatrix}
       1 & 1 & 1 & 0         \\[0.3em]
       1 & 1 & -1 & 0        \\[0.3em]
       1 & -1 & 0 & 1        \\[0.3em]
       1 & -1 & 0 & -1        \\[0.3em]
     \end{pmatrix}
\end{equation*}
}
\normalsize

\noindent From now on, we show a Haar basis matrix by $\mathbf{H}$ whenever the corresponding dimension is understood from the context.

 \begin{definition}[Time-series data \cite{Shumway:2303176}] 
\label{def:time-series}
A time  series $\{x_t\}$,  is  a  sequence  of  observations  on  a  variable  taken  at  discrete  intervals  in  time.  We index the time periods as $1, 2, ..., k$. Given a set of time periods, $\{x_t\}$ is shown  as a column vector $\mathbf{x} = (x_1,\dots, x_k)^T$ or a sequence $x_1x_2\dots x_k$.
\end{definition}

\begin{figure}[h]
    \vspace{-10mm}
	\centering
	\includegraphics[width=.8\linewidth]{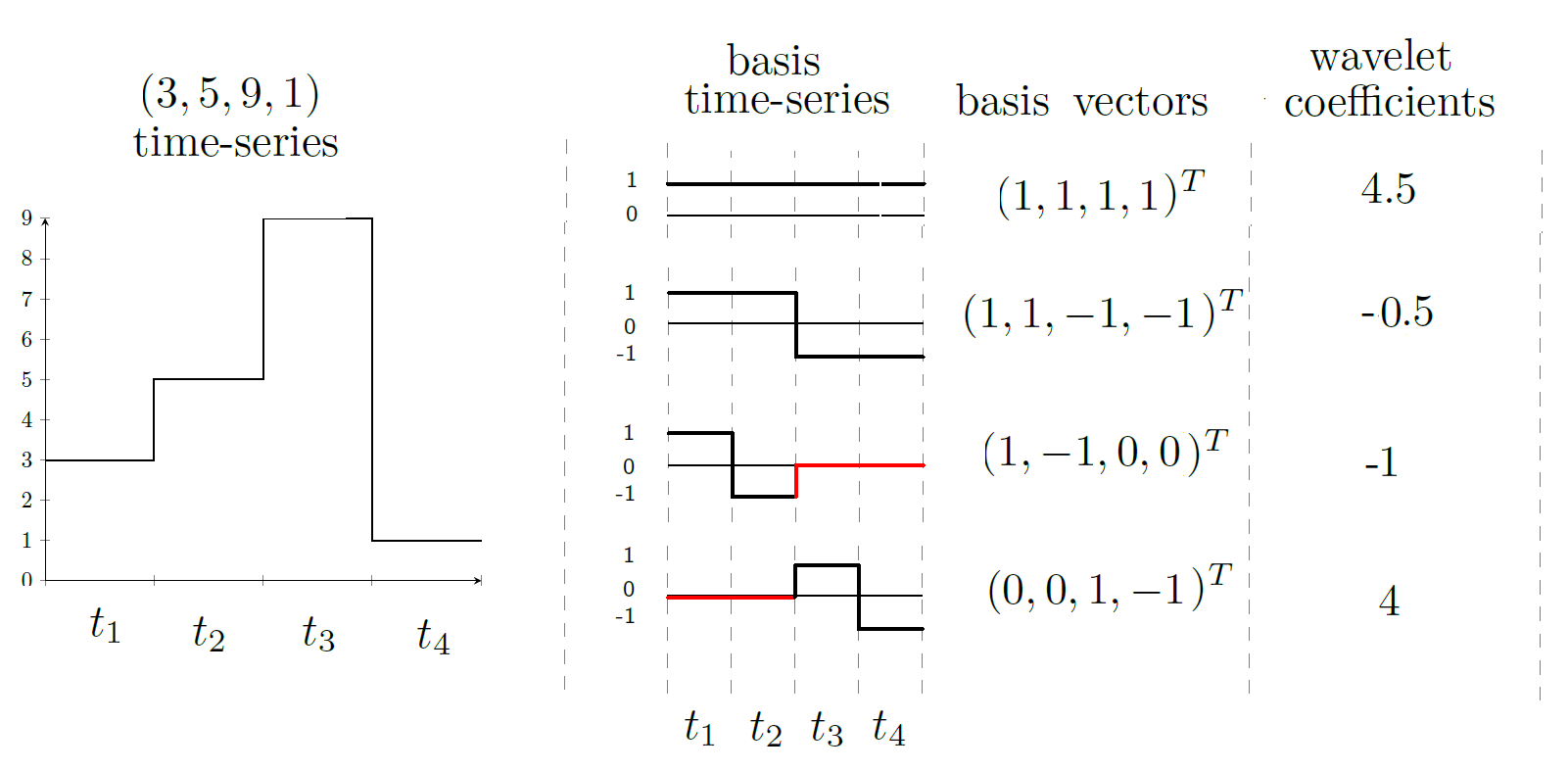}
	\caption{Decomposition of a time-series into a set of Haar Wavelet series}
	\label{fig:Wavelet example}
\end{figure}

Every time-series data can be decomposed into a set of basis time-series called \textit{Haar Wavelet} \cite{Aggarwal15}. A Haar wavelet time-series represents the temporal range of variation in the form of a simple step function. For a given time-series, $\{x_i\}$, of length $n$, the corresponding Haar wavelet basis time-series are shown by  Haar basis vectors in $\mathbb{R}^n$  \cite{GallieLinearAlgebra2019book}, see Def. \ref{def:haar basis vectors}. For example, consider a time-series like $\mathbf{x} = (3,5,9,1)^T$, then it can be decomposed into the sets of Haar wavelet time-series shown in Fig. \ref{fig:Wavelet example}.

In the above example, one sees that each Haar wavelet time-series has a corresponding Haar basis vector. Thus, the input time-series, $\mathbf{x}$, can be represented as the sum of Haar basis vectors with  corresponding coefficients. More compactly, it can be easily represented by the following matrix operation, called \emph{Discrete Wavelet Transformation} (DWT):
\begin{equation}
        \mathbf{x} = \mathbf{H}\mathbf{w} 
    \label{eq: DWT}
\end{equation}
\noindent where $\mathbf{w}$ is a column vector that contains wavelet coefficients. For the above example, Eq. \ref{eq: DWT} is as follow:
\small{
\begin{equation*}
    \begin{pmatrix}
       3           \\[0.3em]
       5           \\[0.3em]
    9               \\[0.3em]
    1              \\[0.3em]
     \end{pmatrix}   =  \begin{pmatrix}
       1 & 1 & 1 & 0         \\[0.3em]
       1 & 1 & -1 & 0        \\[0.3em]
       1 & -1 & 0 & 1        \\[0.3em]
       1 & -1 & 0 & -1        \\[0.3em]
     \end{pmatrix}   \begin{pmatrix}
       4.5           \\[0.3em]
       -0.5           \\[0.3em]
    -1              \\[0.3em]
    4              \\[0.3em]
     \end{pmatrix}
\end{equation*}
}
\normalsize

\vspace{-\baselineskip}

The crucial observation, is that for a given time-series, the set of Wavelet coefficients show available variations in the time-series at different resolutions. The first coefficient is the global average value of time-series, i.e., $w_1 = \frac{3+5+9+1}{4} = 4.5$. The second coefficient shows the difference in average between the average of the first half and the average of the second half, i.e., $\frac{(5+3)/2 - (1+9)/2}{2} = -0.5$. This process can be applied recursively until reaching a single element. Hence, one can use Wavelet coefficients as a sound way to encode time-series that inherit variability information.


\subsection{Feature Generation}
\label{subsec: feature generation}
The technique in this section generates a sets of multidimensional features, $\mathbf{x}$, for every  trace $\sigma$. The procedure can be seen as an encoding task that maps a trace into a multidimensional vector space. This encoding is used to identify both control-flow and performance differences between two process variants. For the sake of simplicity we present it for control-flow dimension. 

The technique in this section provides numerous advantages for analysing sequential data. It is a \emph{lossless} encoding such that the input time-series can be recovered by the Haar basis matrix for that vector space, see Eq. (\ref{eq: DWT}). 
Second, by DWT-encoding the time-series before analysing it,  well-known problems of {\em auto-correlation} and {\em cross-correlation} are significantly alleviated, since the generated features are almost uncorrelated~\cite{Aggarwal15, Bakshi1999MULTISCALEAA}.
Thus, without losing information one can safely work only with wavelet coefficients rather than over the raw data.

Given an input trace, the proposed technique contains three parts, i.e., \emph{binarization}, \emph{vectorization}, and \emph{stacking}. Binarization is a procedure to generate a set of time-series from an input trace. Vectorization encodes every time-series into a vector representation with the help of DWT. Finally, to have a consolidated representation for the generated vectors, they are stacked into a single vector.

The starting point for generating a set of features in a vector space is to represent an input trace as several $\{ 0,1 \}$ time-series that are called binarization. Formally:
\begin{definition}[Binarization]
Given a universal set of activity names $\mathcal{E}$, and trace $\sigma$, function $f()$ maps $\sigma$ into a set of $|\mathcal{E}|$ time-series of length $|\sigma|$, i.e., $\forall e_i \in \mathcal{E}, f: (e_i, \sigma) \rightarrow \{0,1\}^{|\sigma|}$. 
\label{def: trace binarization}
\end{definition}

\noindent The above definition provides time-series of zeros for an event that does not exist in an input trace. 
This way one can represent all traces in a unique vector space \footnote{In practice we keep only non-zero elements in the implementation.}. For example, consider a trace like $\sigma = e_1e_2e_1e_1$, with $\mathcal{E} =\{e_1,e_2, e_3\}$, then, $f(e_1, \sigma) = 1011$, $f(e_2, \sigma) = 0100$, and $f(e_3, \sigma) = 0000$.

Binarization of a given trace provides a time-series for each event of it. Vectorization based on DWT, see Eq. \ref{eq: DWT}, captures simultaneously frequency and location in time information, and embeds auto-correlation and cross-correlation information in the generated features. Formally:


\begin{definition}[Vectorization]
 Given a time-series $\mathbf{x} = (x_1,x_2,\dots,x_n)^T$ where $x_i \in \{0,1\}$, function $g()$,  computes the corresponding wavelet coefficients, $\mathbf{w}$, for $\mathbf{x}$, i.e., $ g(\mathbf{x}) = \mathbf{w}= \mathbf{H}^{-1}\mathbf{x}$.
\label{def: trace vectorization}
\end{definition}
 \noindent In the above definition, $\mathbf{H}^{-1}$ is the inverse of Haar basis matrix for $\mathbb{R}^n$. For example, for time-series $\mathbf{x}^{(1)} = (1,0,1,1)^T$, $\mathbf{x}^{(2)} = (0,1,0,0)^T$, and  $\mathbf{x}^{(3)} = (0,0,0,0)^T$ the corresponding wavelet coefficients $\mathbf{w}^{(1)}$,$\mathbf{w}^{(2)}$, and $\mathbf{w}^{(3)}$ are as follows:

\small{
\begin{equation*}
    \underbrace{\begin{pmatrix}
       0.75           \\[0.3em]
       -0.25           \\[0.3em]
    0.5               \\[0.3em]
    0              \\[0.3em]
     \end{pmatrix}}_\text{$\mathbf{w}^{(1)}$}   =  \underbrace{\begin{pmatrix}
       0.25 & 0.25 & 0.25 & 0.25        \\[0.3em]
       0.25 & 0.25 & -0.25 & -0.25        \\[0.3em]
       0.5 & -0.5 & 0 & 0        \\[0.3em]
       0 & 0 & 0.5 & -0.5        \\[0.3em]
     \end{pmatrix}}_\text{$\mathbf{H}^{-1}$}  \underbrace{ \begin{pmatrix}
       1           \\[0.3em]
       0           \\[0.3em]
    1              \\[0.3em]
    1              \\[0.3em]
     \end{pmatrix}}_\text{$\mathbf{x}^{(1)}$}, \quad \underbrace{\begin{pmatrix}
       0.25           \\[0.3em]
       -0.25           \\[0.3em]
    -0.5               \\[0.3em]
    0              \\[0.3em]
     \end{pmatrix}}_\text{$\mathbf{w}^{(2)}$},  \quad \underbrace{\begin{pmatrix}
       0           \\[0.3em]
       0           \\[0.3em]
    0              \\[0.3em]
    0              \\[0.3em]
     \end{pmatrix}}_\text{$\mathbf{w}^{(3)}$}
\end{equation*}
}

\normalsize

\noindent According to the above example, for a trace like $\sigma = e_1e_2e_1e_1$, to have a consolidate representation we stack together the coefficient vectors into a single vector, formally:


\begin{definition}[Vector stacking]
    For an input trace $\sigma$, and universal of activities $\mathcal{E}$, with $|\mathcal{E}| =k$, lets assume that $\mathbf{w}^{(1)}, \mathbf{w}^{(2)}, \dots,\mathbf{w}^{(k)}$ show the corresponding wavelet coefficients vectors, then   the stacked vector is defined as \small{$\mathbf{w}^{(\sigma)} = (\mathbf{w}^{(1)^T}, \mathbf{w}^{(2)^T}, \dots, \mathbf{w}^{(k)^T})$}.
\label{def: vector stacking}
\vspace{-3mm}
\end{definition}

\noindent Regarding to the above definition,  a design matrix $\mathbf{D}$ for a process variant is defined to be a matrix whose rows are the stacked vectors of the corresponding traces.  As an example for a process variant containing only two traces, i.e., $\sigma_1 = e_1e_2e_1e_1 $, and $\sigma_2 = e_1e_2e_3e_1$, the respective design matrix after binarization, vectorization, and stacking is as follow:

\small{
\[
\mathbf{D} = \bordermatrix { ~ & e_1   & e_1  & e_1    &  e_1   &  e_2   & e_2    & e_2  &  e_2   & e_3 & e_3 & e_3 & e_3 \cr
\mathbf{w}^{(\sigma_1)} & 0.75   & -0.25   & 0.5    &   0  &  0.25   & -0.25    & -0.5  &    0  & 0  & 0 &0 &0 \cr
\mathbf{w}^{(\sigma_2)} & 0.5   &  -0.25   & 0      &   0  &   0.25   & -0.25     & -0.5  &    0  & 0.25  & -0.25 & 0 & 0.5 \cr}
\]
}
\normalsize

\noindent One can see that the first four columns show the wavelet coefficients for event $e_1$, the second four columns show the wavelet coefficients for event $e_2$, and so on. It is easy to see that wavelet coefficients for $e_2$ are the same for $\sigma_1$ and $\sigma_2$; however for $e_1$ only three out of four coefficients are equal which shows different frequency and location of this event between $\sigma_1$ and $\sigma_2$. 

    In practice, to have a unique and common vector space for the encoding of two process variants, we set the dimension of the vector space to the length of the longest trace for both variants. If the alphabet of process variants are different then the union of them is considered as the universal alphabet. Also, an analyst can generate different kind of features; for example one can create a design matrix for adjacent events in traces, where the features are like $e_ie_j$, with $j= i+1$, instead of only single events.
    
    \textbf{Time complexity}. The time complexity of the proposed approach  is cubic on the length of the longest trace in the worst-case. However, in practice it is much less than this amount: lets assume that there are two process variants $\varsigma_1$, $\varsigma_2$ with $n_1$, $n_2$ number of traces respectively, $\mathcal{E}$, is the universal activity names, and $d=\text{max}|\sigma|, \forall \sigma \in \varsigma_1, \varsigma_2$, is the length of the longest trace between variants. Thus, computing the Haar basis matrix and its inverse for $\mathbb{R}^d$ require $\mathcal{O}(log_2d)$, and $\mathcal{O}(d^3)$\footnote{Note that the cubic complexity is the required time for computing the inverse matrix from scratch. To this end, there are much more efficient approaches like Coppersmith–Winograd algorithm with $\mathcal{O}(d^{2.37})$.} operations respectively. It must be mentioned that \cite{Yi2000} proposed $\mathcal{O}(d^2)$ for computing the inverse of a matrix in an incremental way. To create the design matrix $\mathbf{D}^{(i)}, \text{ for } i=1,2$, the number of required operations is $\mathcal{O}(n_i*(d*\mathcal{E}))$. However, this matrix is very sparse since for an input trace $\sigma$, only the entries related to $e_i \in \sigma$ are non-zero. Another possibility to alleviate significantly the overall complexity is by precomputing and storing Haar matrices.

\subsection{Feature Selection}
\label{subsec: feature-selection}
This section presents a novel feature selection method, grounded on machine learning,  that captures the statistically significant features between two design matrices, i.e., $\mathbf{D}^{(i)}, \text{ for } i=1,2$.  Generally speaking, the representation of an entity in an arbitrary vector space contains a set of \emph{relevant} and \emph{irrelevant} features. Though, it is unknown as prior knowledge. Thus, a feature selection algorithm  is the process of selecting a subset of relevant features that are the most informative ones with respect to the class label. 

Though feature selection procedures have numerous advantages for machine learning algorithms, in this paper, we leverage the idea of feature selection to highlight the existing differences between two process variants (class 1 and class 2). 
It must be stressed that every events $e_i \in \mathcal{E}$ is represented by a set of features (columns) in the designed matrices, see Def. \ref{def: vector stacking}, and each row is called an instance.
The feature selection technique in this paper is a \emph{wrapper} method, where a classifier is trained on a subset of features. If the trained classifier provides acceptable performance according to some criteria for unseen instances, i.e., test instances, then the subset of features is selected and called \emph{discriminatory} features.

Before proceeding, and for the sake of exposition we stack the design matrices $\mathbf{D}^{(1)}, \mathbf{D}^{(2)}$, with the corresponding class labels into a matrix called \emph{augmented design matrix} as follow:
\small{
\begin{equation}
\mathbf{X} = \left(\begin{array}{@{}c|c@{}}
  \mathbf{D}^{(1)}
  & \mathbf{1} \\
  
\mathbf{D}^{(2)} & \mathbf{2}
\end{array}\right)
\label{eq:augmented matrix}
\end{equation}
}
\normalsize
\noindent Where $\mathbf{1}, \mathbf{2}$ are column vectors showing the class labels. It is clear that $\mathbf{X}$ has $d \times \mathcal{E} +1$ features or columns. From $\mathbf{X}$, and for a subset of features, $\mathcal{S} \subseteq \mathcal{E}$, we split $\mathbf{X}$ into training and test datasets and denote them by $\mathbf{X}_{:,\mathcal{S}}^{(train)}$, and $\mathbf{X}_{:,\mathcal{S}}^{(test)}$. It must be mentioned that, to create training and test datasets we use \emph{stratified sampling} method which keeps the proportion of classes similar in either datasets \cite{Liu2007ComputationalMO}. Stratified sampling helps to create sub-samples that are representative of the initial dataset.

\begin{definition}[Discriminatory feature] 
A subset of features, $\mathcal{S}$, with $|\mathcal{S}| \leq d \times \mathcal{E}+1$ is discriminatory if a  binary classification function $f: \mathbb{R}^{\mathcal{|S|}} \rightarrow \{1,2\}$ that is trained on $\mathbf{X}_{:,\mathcal{S}}^{(train)}$, provides acceptable performance according to some criteria for unseen instances in $\mathbf{X}_{:,\mathcal{S}}^{(test)}$.
\label{def: discriminatory feature}
\end{definition}


\noindent Definition \ref{def: discriminatory feature} does not pose any  restrictions on the shape or the number of parameters for $f()$, indeed, according to \textit{universal approximation theorem} there exist such a mapping function between any two finite-dimensional vector spaces given enough data \cite{HORNIK1989359}.

 There are several ways to measure the performance of a classifier on unseen instances. An appropriate metric for imbalanced classes is $F_1$ score. It  measures the performance of a classifier for one of classes only, e.g., class 1 or positive class, and does not take into account the true negatives, i.e., correct predictions for the other class into account, hence some information are missed \cite{Powers2011EvaluationFP}. Since in our setting two classes (i.e., process variants) are equally important, and for each subset of features the proportion of class labels varies, and probably imbalanced, we propose \emph{weighted} $F_1$ score as follow:
\begin{equation}
    \Bar{F_1} = \gamma_1 F^{(1)}_1 +\gamma_2 F^{(2)}_1
\label{eq:weighted f-score}
\end{equation}

\noindent Where $F^{(i)}_1, \text{ for } i=1,2$ shows the $F_1$ obtained by the classifier for classes (i.e., 1, 2) on $\mathbf{X}_{:,\mathcal{S}}^{(test)}$. The coefficients  $\gamma_1$, $\gamma_2$ shows the proportion of class labels in the test dataset. It must be noted that the values of $\gamma_1$ and $\gamma_2$ varies for different subset of features. For example, assume that $\mathbf{X}^{(test)}$ contains three instances as shown bellow:

\begin{figure}
\vspace{-5mm}
    \centering
    \includegraphics[width=1\linewidth]{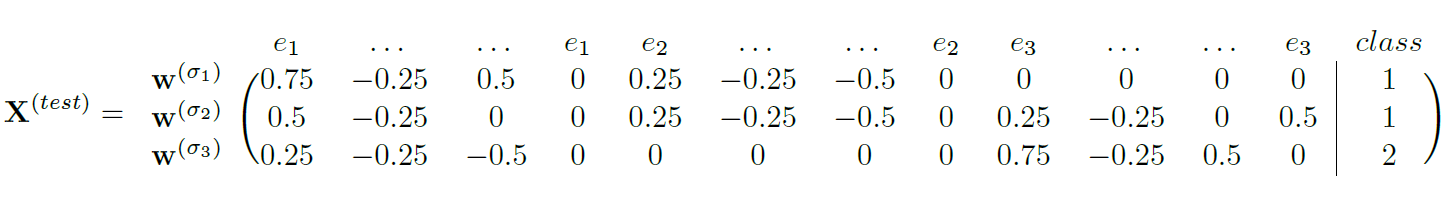}
    \vspace{-10mm}
\end{figure}

\noindent The test dataset corresponds to the wavelet coefficients for three traces $\sigma_1 = e_1e_2e_1e_1 $,  $\sigma_2 = e_1e_2e_3e_1$, and $\sigma_3 = e_3e_1e_3e_3$, where the first two traces belong to one process variant (class 1), and the last trace comes from another process variant (class 2).  One can see that if we consider $\mathcal{S}$ as the columns related to $e_1$, then the proportion of classes are $\frac{2}{3}$ and $\frac{1}{3}$, however for columns related to $e_3$ both numbers are $\frac{1}{2}$. The reason is, an arbitrary trace $\sigma_i$ contains portion of alphabet $\mathcal{E}$, hence the coefficients $\gamma$ vary from subset to subset, and must be adjusted dynamically.

    

For a subset of features, $\mathcal{S}$, the worst performance of a classifier, $f()$, happens when it provides a single label for all test instances. It takes place when there is not enough information in $\mathcal{S}$ for the classifier to discriminate classes (process variants). In more details, if $n_1$, and $n_2$ show the number of instances for each class in $\mathbf{X}_{:,\mathcal{S}}^{(test)}$, then the worst performance happens when $f()$ labels all test instances as class 1 ( $F^{(1)}_1 = \frac{2n_1}{n_2 + 2n_1}$ ), or as class 2 ( $F^{(2)}_1 = \frac{2n_2}{n_1 + 2n_2}$ ), therefore we denote the worst performance of classifier $f()$ by $\Bar{F_1}^{(0)}$, that is defined as the weighted average of worst cases as follows:
\small{
\begin{equation}
  \Bar{F_1}^{(0)} = \underbrace{\frac{n_1}{n_1 + n_2}}_{\gamma_1} \times \underbrace{\frac{2n_1}{n_2 + 2n_1}}_{F^{(1)}_1}    +    \underbrace{\frac{n_2}{n_1 + n_2}}_{\gamma_2} \times \underbrace{\frac{2n_2}{n_1 + 2n_2}}_{F^{(2)}_1}  
  \label{eq: worst performance}
\end{equation}
}
\normalsize

\noindent One must note that the value of $\Bar{F_1}^{(0)}$ like coefficients $\gamma_1$ and $\gamma_2$ varies for different $\mathcal{S}$; thus it is adjusted dynamically.

Regarding Eq. \ref{eq: worst performance}, for a subset of features, $\mathcal{S}$, we define acceptable performance for a classification function $f()$  if its performance measured by $\Bar{F_1}$ score, is statistically greater than the corresponding $\Bar{F_1}^{(0)}$ score at some significant level $\alpha$. Formally, we formulate a statistical test containing the following hypotheses:
\small{
\begin{equation}
    \underbrace{H_0: \Bar{F_1} = \Bar{F_1}^{(0)}}_\text{Null-hypothesis}, \quad
    \underbrace{H_1: \Bar{F_1} > \Bar{F_1}^{(0)}}_\text{Alternative-hypothesis}
    \label{eq:hypothesis testing}
\end{equation}
}
\normalsize

\noindent The null-hypothesis ($H_0$) in Eq. \ref{eq:hypothesis testing} assumes that for a subset of features $\mathcal{S}$, the classifier $f()$ is unable to discriminate test instances; in other words if $\mathcal{S}$ represents columns relating to a set of events, then, it claims that the control-flows containing these events are not statistically different between process variants, whereas the alternative-hypothesis ($H_1$) claims they differ. 

To make the statistical test in Eq. \ref{eq:hypothesis testing} work, we invoke \emph{stratified k-fold cross-validation} procedure to measure the performance of the classifier $k$ times for $\mathcal{S}$. In each round of cross-validation, different training and test datasets, $\mathbf{X}_{:,\mathcal{S}}^{(train)}$, and $\mathbf{X}_{:,\mathcal{S}}^{(test)}$ are obtained via stratified sampling method, and the corresponding $\Bar{F_1}$ score is calculated. Based on \emph{Central Limit Theorem} (CLT), the average of $\Bar{F_1}$ scores ($k$ times) approximates a normal distribution for $k>30$ or a $t-$distribution for small $k$ numbers \cite{wackerly08}.

\textbf{Complexity}. In practice, a feature selection algorithm has to do an exhaustive search to obtain subsets of features that are representative of the input dataset. In general, given $n$ features there could be $2^n$ candidates for selecting subset of features; however in our setting, the search space is limited to subsets of adjacent features 
available in process variants. Therefore the respective search space reduces drastically. 


\subsection{Filtering}
\label{subsec: filtering}

This section elucidates the findings of the previous step. In fact, identifying discriminatory events, though statistically significant, does not provide enough insights for the analyst to understand the existing differences. To bring this information into a human-readable form, one can create a  directly-follows graph for each process variant by only considering those traces that carry information about discriminatory parts. Technically, assume $\mathcal{S}$ contains features relating to event $e_i$, and it was found to be statistically significant between two process variants, then all traces containing $e_i$ are kept. This procedure continues for all discriminatory elements (an event or a set of them). A \emph{mutual fingerprint} is a  directly-follows graph created based on these sets of traces for each process variant separately.


\vspace{-2mm}

\section{Evaluation}
\label{sec:Experiment}
We implemented our approach in Python 2.7 and used this prototype tool to evaluate the approach over different real-life datasets, against two baselines \cite{bolt2018, Nguyen2018-Multiperspectquteprints117962}. As discussed in Section \ref{subsec: feature-selection}, the proposed feature selection approach can be coupled with any classifier. We trained a Support Vector Machine (SVM) with Radial Basis Function (RBF) kernel and ten times stratified cross validation of the results. We used SVM with RBF because it has been shown that this machine learning method deals well with sparse design matrices \cite{Aggarwal15}, like those that we build for the process variants in our datasets (a great portion of entries in these matrices are zero). The experiments were conducted on a machine with an Intel Core i7 CPU, 16GB of RAM and MS Windows 10. 

\subsection{Setup and Datasets}
Table \ref{table:dataset and variants} provides descriptive statistics for the four real-life event logs that we used in our experiments. We obtained these  datasets from the ``4TU Data Center'' public repository \cite{4dtu}. The logs cover different processes: road traffic fine management process at an Italian municipality (RTFM log), flow of patients affected by sepsis infection (SEPSIS), IT incident and problem management at Volvo (BPIC13), and permit request at different Dutch municipalities (BPIC15). For each log, we established two process variants on the basis of the value of an attribute (e.g. in the case of the RTFM log, this is the amount of the fine, while in the case of SEPSIS this is the patient's age), in line with \cite{Nguyen2018-Multiperspectquteprints117962}. The attributes used to determine the variants, and their values, are also reported in Table \ref{table:dataset and variants}. As we can see from the table, each log has class imbalance (one variant is much more frequent than the other).
Due to lack of space, in the rest of this section we focus on the RFTM log only. The interested reader can find the results for all four logs online.\footnote{\url{https://doi.org/10.6084/m9.figshare.10732556.v1}} 



\begin{table}[h]
\vspace{-9mm}
	\caption{Datasets and corresponding process variants \cite{4dtu}}
	\label{table:dataset and variants}
	\centering 
	\small \begin{tabular}{|p{1.5cm}| p{3.0cm}| p{2.0cm} | p{1cm} | p{1cm} | p{1cm}| p{2.0cm}|}
		\hline
     \multicolumn{7}{|c|} {Event log}\\
		\hline \hline
		Event log &Process Variant  & Cases (uni.) & $|\sigma|_{min}$ & $|\sigma|_{max}$ & $|\sigma|_{avg}$ & $|Events|$ (uni.)\\ [1ex]
		\hline
	 	\multirow{2}{*}{RTFM} & 1) Fine's amount $\geq$ 50  & 21243 (159) &    2   & 20 & 4  &91499 (11) \\
		  & 2) Fine's amount $<$50          &  129127 (169)  & 2 & 11 & 4 &  469971 (11)\\
		\hline
		\multirow{2}{*}{SEPSIS} & 1) Patient's age $\geq$ 70  & 678 (581) &    3   & 185 & 15  &10243 (16) \\
		  & 2) Patient's age $\leq$35          &  76 (51)  & 3 & 52 & 9 &  701 (12)\\
		\hline
		\multirow{2}{*}{BPIC13} & 1) Organization = $A_2$  & 553 (141) &   2   & 53 & 8  &4221 (3) \\
		  & 2)  Organization = $C$        &  4417 (611)  & 1 & 50  & 7 &  29122 (4)\\
		\hline
		
		\multirow{2}{*}{BPIC15} & 1) Municipality = 1  & 1199 (1170) &   2   & 62 & 33.1  &36705 (146) \\
		  & 2)  Municipality = 2       &  831 (828)  & 1 & 96  & 38.6 &  32017 (134)\\
		\hline
		
	\end{tabular}
	\vspace{-5mm}
\end{table}

\normalsize

\begin{figure}[h]
\vspace{-5mm}
	\centering
	\includegraphics[width=1.1\linewidth]{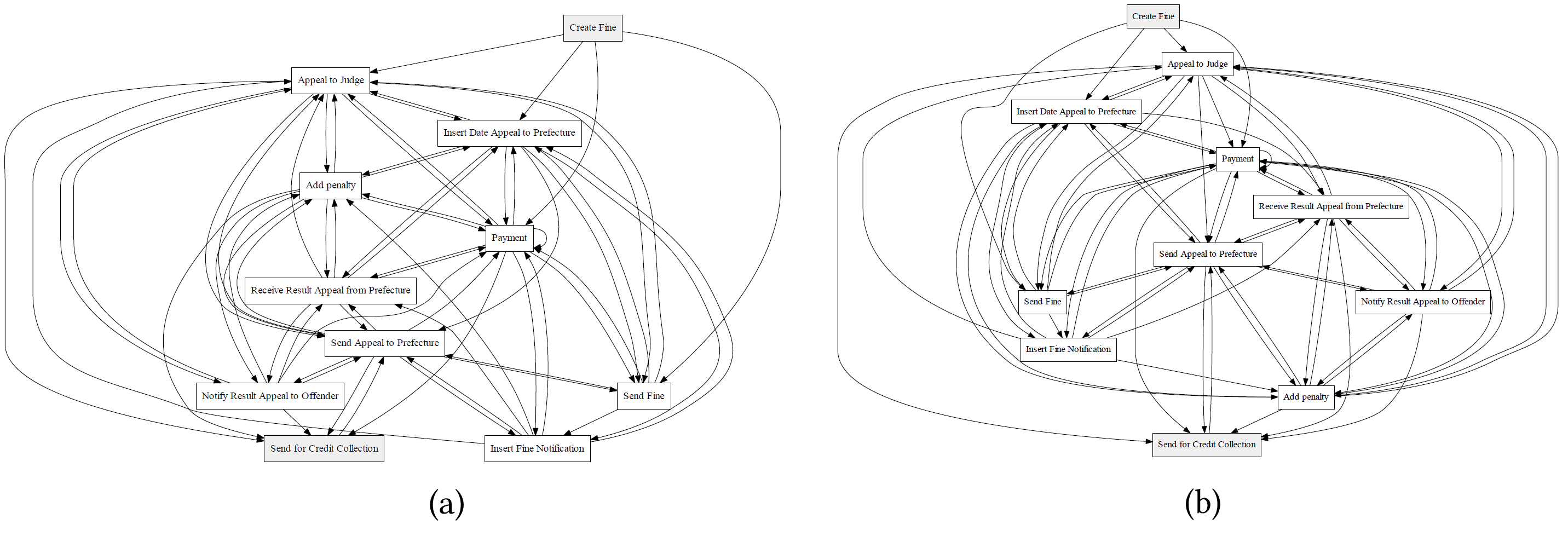}
	\vspace{-7mm}
	\caption{\small{Directly-follows graph of the two process variants of the RFTM log: (a) Fine's amount $\geq$ 50; (b) Fine's amount $<$ 50}}
	\label{fig:fine road variants}
	\vspace{-3mm}
\end{figure}

Figure \ref{fig:fine road variants} shows the directly-follows graph (a.k.a.\ process map) for the two process variants of the RTFM log: the first one, obtained from the sublog relative to fines greater than or equal to 50 EUR, the other obtained from the sublog relative to fines lower than 50 EUR. Process maps are the common output of automated process discovery techniques. To aid the comparison between process variants, the arcs of process maps can be enhanced with frequency or duration statistics, such as case frequency or  average duration. Yet, even with such enhancements, when the graphs are dense like those in Figure \ref{fig:fine road variants}, identifying control-flow or duration differences between the two variants becomes impracticable. Accordingly, the main objective of our evaluation is to use our approach and the two baselines to answer the following research question:

\begin{itemize}
    \item \textit{RQ1: What are the key differences in the control flow of the two process variants?}
\end{itemize}

After we have identified these key differences, we can enrich the results with an analysis of the differences in activity duration, leading to our second research question:

\begin{itemize}
    \item \textit{RQ2: What are the key differences in the activity durations of the two process variants?}
\end{itemize}

To answer RQ1, we apply our approach by considering which pair of adjacent events, called an edge, i.e., $e_ie_{i+1}$ is selected as a discriminatory edge. We consider only edges that are available in process variants. An edge shows the finest control-flow unit. Next, to answer RQ2, we include an additional analysis 
to examine whether the average duration time for an edge varies significantly between process variants. Essentially RQ2 boils down to running a statistical test between the corresponding average duration times of the same edge in the two variants.

\subsection{Results}
\label{subsec:result}


Table \ref{table, road teraffic significant part} shows the edges in the two directly-follows graphs of Fig. \ref{fig:fine road variants} that are statistically significantly different both in frequency and in time (i.e.\ the temporal order of execution within a path of the graph), as obtained by our approach. 

\begin{table}[h]
	\caption{Significantly-different edges in frequency and order, between the directly-follows graphs of Fig. \ref{fig:fine road variants}, obtained with our approach}
	\centering 
	\small \begin{tabular}{| p{5.9cm}| p{1.2cm} | p{1.2cm} | p{1cm} | p{1cm}| p{1.2cm}|}
		\hline
		Edge  & $\gamma_1$& $\gamma_2$ &  $\bar{F_1}^{0}$ & $\bar{F_1}$ & P-value \\ [1ex]
		\hline
		('Add penalty', 'Payment')  & 0.17  &  0.83  &  0.18  &    0.35  & 0.009  \\
		\hline
		('Payment', 'Payment')     &  0.19  &   0.81   &  0.20  & 0.38    & 0.006  \\
		\hline
		('Payment', 'Send for Credit Collection') &  0.23  &  0.77  & 0.22 & 0.34 & 0.006\\
		\hline
	\end{tabular}
		\label{table, road teraffic significant part}
		\vspace{-5mm}
\end{table}

Table \ref{table, road teraffic significant part} also contains the classifier's performance, measured by $\bar{F}_1$. For each edge, this score is statistically greater (averaged from ten times cross-validation) than the corresponding worst case, $\bar{F_1}^{0}$, see Eq. \ref{eq: worst performance}. Besides, we note that the coefficients $\gamma_1$ and $\gamma_2$ vary for each edge. This shows that the proportion of class labels vary for each edge. The two baselines \cite{bolt2018, Nguyen2018-Multiperspectquteprints117962} cannot provide such results. It is because they only consider the relative frequency of an edge in each process variant, and apply a statistical test on such frequencies, and neglect the \textit{order} in which such edge occurs within a path of the directly-follows graph. 
Thereby, they miss to capture the backward and forward information in a trace (or path in the directly-follows graph) containing that edge. 

\begin{figure}[h]
	\centering
	\includegraphics[width=.65\linewidth]{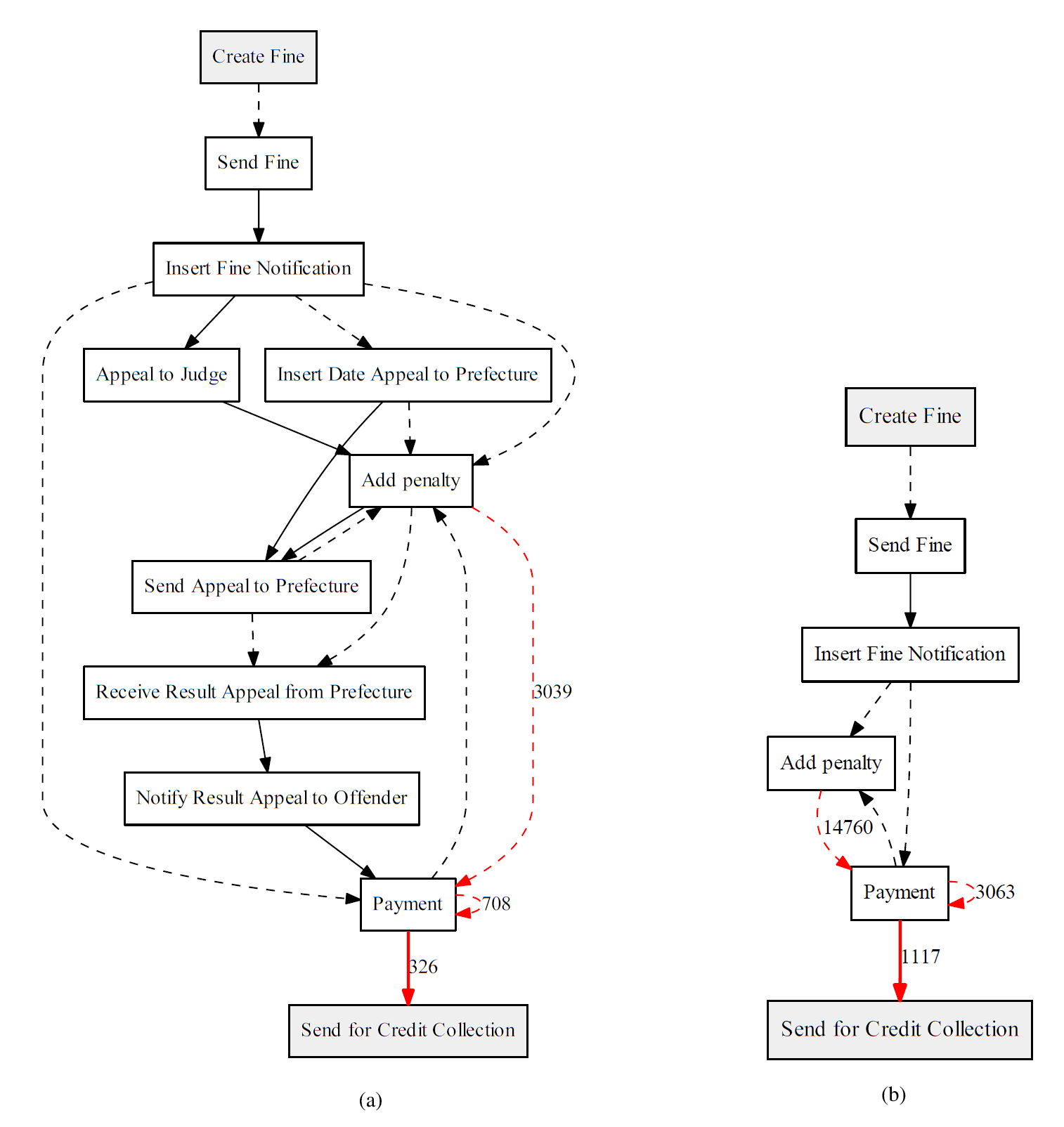}
	\caption{\small{ Answer to RQ1: Discovered mutual fingerprints for the RFTM variants in Table \ref{table:dataset and variants}; (a) Fingerprint for variant Fine's amount $\geq$ 50, (b) Fingerprint for variant Fine's amount $<$ 50}}
	\label{fig:fine road fingerprint}
\end{figure}

\begin{figure}[h]
	\centering
	\includegraphics[width=1\linewidth]{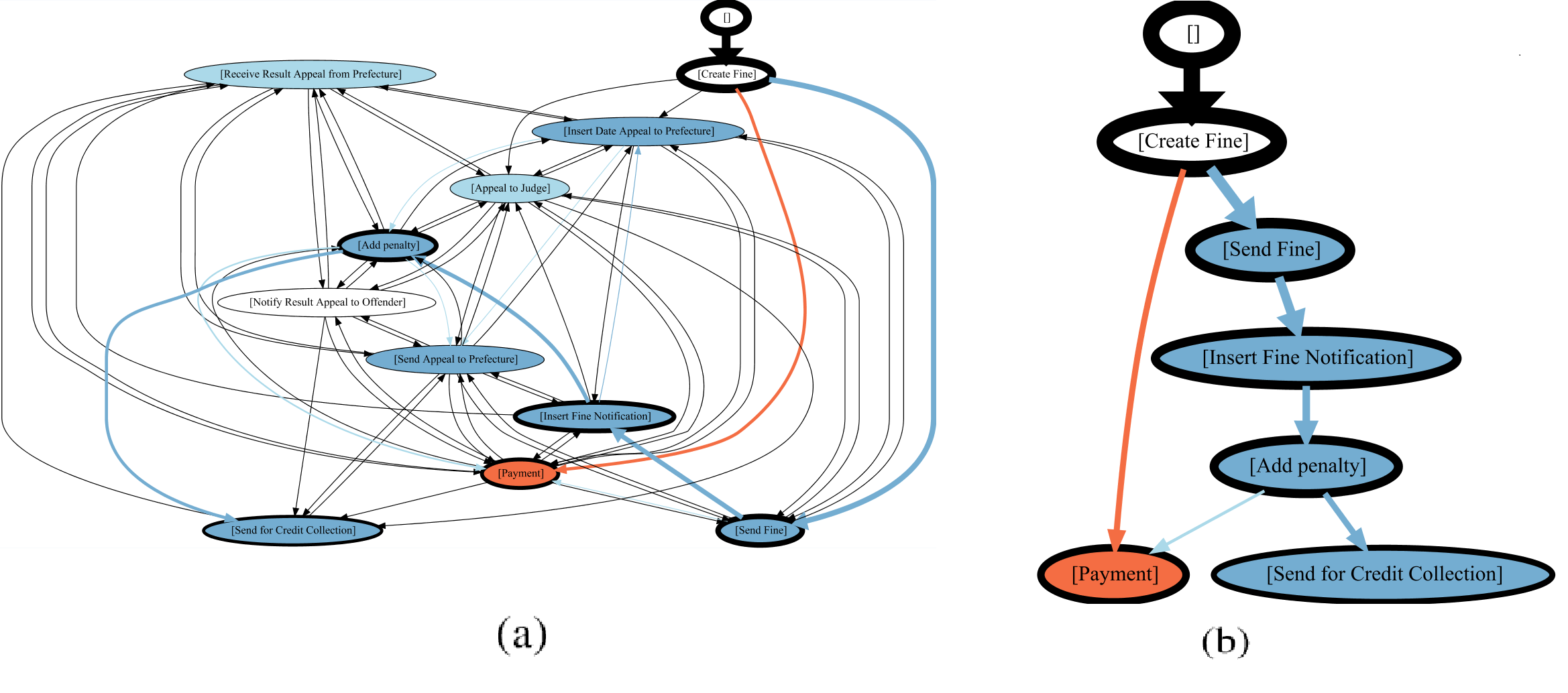}
	\vspace{-3mm}
	\caption{\small {Answer to RQ1: Directly-follows graph obtained for the two variants in the RFTM log according to \cite{bolt2018}. The frequency of an edge is shown by its thickness while the color shows the difference between the edge frequencies in the two variants: (a) all edges shown; (b) edges with frequency $\leq \%5$ are cut}}
	\label{fig:fine road Bolt}
	\vspace{-6mm}
\end{figure}

Figure \ref{fig:fine road fingerprint} shows the mutual fingerprints resulting from the edges in Table \ref{table, road teraffic significant part}. For comparison purposes, Figure \ref{fig:fine road Bolt} shows two annotated directly-follows graphs obtained with the baseline in \cite{bolt2018}.\footnote{\footnotesize{Obtained using the default settings in ProM 6.9}}

For ease of discussion, let us neglect the type of edge (solid, dashed) in Fig. \ref{fig:fine road fingerprint}. The edges in Table \ref{table, road teraffic significant part} are highlighted in red in Fig. \ref{fig:fine road fingerprint}. 
In effect, traces that contain a discriminatory edge, like \emph{(Payment, Send for Credit Collection)} differ between process variants. An offender whose fine's amount is greater than or equal to 50 Euros goes through more steps, i.e.\ through a different control flow as shown in Fig. \ref{fig:fine road fingerprint} (a). In contrast, an offender whose fine's amount is below 50 Euros, goes to less steps, as in Fig. \ref{fig:fine road fingerprint} (b). 

In contrast, the baseline in \cite{bolt2018} comes up with a single directly-follows graph for both process variants (Fig. \ref{fig:fine road Bolt}). We note that the approach in \cite{Nguyen2018-Multiperspectquteprints117962} produces similar results, which we did not show due to lack of space. The baselines are unable to identify control-flow differences completely. Rather, they show (statistically-significant) differences at the level of individual edges. For example, in \cite{bolt2018}, the thickness of each edge shows the frequency of that edge. Even if a statistical test is applied for each edge to determine whether the corresponding frequency varies between process variants, this information is not sufficient to identify differences in paths. The problem is exacerbated by the fact that a directly-follows graph generalises process behavior since the combination of the various edges give rise to more paths than the traces in the event log. Indeed the baselines \cite{bolt2018, Nguyen2018-Multiperspectquteprints117962} consider only the edge's frequency, whereas our approach considers both frequency and location of an edge simultaneously.
For example, the approach in \cite{bolt2018} identifies that the frequency of \emph{(Create Fine, Payment)} (the orange edge) is different between process variants. In constrast, in our approach it was found that this particular edge does not discriminate the two process variants from a control-flow perspective. In fact, the paths containing this edge, though having different frequencies, are very similar in the two variants.

Also, "Insert fine notification $\rightarrow$ Appeal to Judge" is not depicted in Fig. \ref{fig:fine road fingerprint} (b), since it does not contribute to any statistically-significant control-flow difference between the two variants. However, it appears in Fig. \ref{fig:fine road fingerprint} (a) because it is in any path that contains at least one of the edges in Table \ref{table, road teraffic significant part}. 
This is a good feature of our approach,
since the edge in question itself does not contribute to any differences, but its occurrence and location affect other edges for the respective variant, giving rise to  different mutual fingerprints. That said, as a side effect, sometimes the resulting fingerprints might contain a large number of edges. In contrast,
the baselines \cite{bolt2018, Nguyen2018-Multiperspectquteprints117962} are unable to capture such correlation because the frequency of an edge is considered in their analysis.



\begin{table}[h]
	\caption{Answer to RQ2: Comparing edge durations between the two variants}
	\centering 
	\footnotesize \begin{tabular}{| p{7.3cm}| p{1.5cm} | p{1.5cm} | p{1cm} | p{1cm}| p{1.5cm}|}
		\hline
		Edge  & $\overline{\Delta t}_1$ (day) & $\overline{\Delta t}_2$ (day) &  P-value \\ [1ex]
		\hline
		('Create Fine', 'Send Fine') & 72.48  & 92.98 & 0  \\
		\hline
		('Create Fine', 'Payment') & 11.52  & 10.38 & 0.0089  \\
		\hline
		('Send Appeal to Prefecture', 'Add penalty')  & 26.13  & 20.00 & 0.0004  \\
		\hline
		('Add penalty', 'Payment') & 152.38  & 169.43 & 0  \\
		\hline
		('Add penalty', 'Receive Result Appeal from Prefecture') & 58.17  & 46.39 & 0.0151  \\
		\hline
		('Payment', 'Payment') & 77.60  & 101.97 & 0  \\
		\hline
		('Payment', 'Add penalty') & 30.94  & 33.27 & 0.0086  \\
		\hline
			('Insert Fine Notification', 'Payment') & 28.83  & 26.48 & 0.0083  \\
		\hline
		('Insert Fine Notification', 'Insert Date Appeal to Prefecture') & 34.24  & 35.50 & 0.0069  \\
		\hline
		('Insert Date Appeal to Prefecture', 'Add penalty') & 22.94  & 24.96 & 0.0016  \\
		\hline
		
		('Send Appeal to Prefecture', 'Receive Result Appeal from Prefecture') & 49.25  & 56.19 & 0.0354  \\
		\hline
	\end{tabular}
		\label{table, road teraffic duration time}
		\vspace{-5mm}
\end{table}
\normalsize

To answer RQ2, we compared the average duration time $\Delta t$ for each edge of the two process variants (capturing the activity duration), and then applied $t$-tests with unequal variances. The results are shown in Table \ref{table, road teraffic duration time}. We superimposed those edges with statistically significant differences in duration, over the corresponding fingerprints by using dashed edges, as shown in Fig. \ref{fig:fine road fingerprint}. 
Both baseline \cite{bolt2018, Nguyen2018-Multiperspectquteprints117962} provide the same results for the duration time. 

\begin{table}[h]
	\caption{Performance of the proposed approach}
	\centering 
	\footnotesize \begin{tabular}{| p{2.3cm}| p{2.5cm} | p{4.0cm} | p{1.8cm} | p{1cm}| p{1.2cm}|}
		\hline
		Dataset  & Execution time (s)& Memory usage (MB) $95\%$ C.I. \\ [1ex]
		\hline
		RTFM & 2340  &  (473 - 608) \\
		SEPSIS & 217  &  (170 - 218)  \\
		BPIC13 & 152  &   (380 - 410) \\
		BPIC15 & 3470  &   (980 - 1040) \\
		\hline
	\end{tabular}
		\label{table:exetime}
		\vspace{-6mm}
\end{table}
\normalsize

\paragraph{Execution time} 
Table \ref{table:exetime} shows the execution time of our approach for each dataset. Time performance is affected by the size and complexity of the event log. For example, in the table we can observe longer times for the RFTM log (39 min), where the number of cases is high, and for BPIC15 (58 min), where the number of unique events is relatively high. Yet, the approach performs within reasonable bounds on a standard laptop configuration. Comparatively, the two baseline techniques are much faster, namely in the order of a few minutes.

Table \ref{table:exetime} also shows RAM occupancy. This is monitored every 10 seconds, and next, the 95\% confidence interval is computed. One can see that the amount of memory for BPIC15 is larger than the other datasets. This can be attributed to many unique events available in each process variant, which give rise to an augmented matrix with a high number of columns. Yet, memory use is quite reasonable (ranging from 473MB min for RFTM to 1.04GB max for BPIC15).

\section{Conclusion}
\label{sec:conclusion}
In this paper, we presented a novel approach to identify statistically-significant differences in the control-flow and activity durations of two business process variants, each captured as an event log. 
The cornerstone technique of this approach is the construction of a novel encoding of discriminatory features from an event log, called mutual fingerprint, based on a discrete wavelet transformation of time series extracted from the traces of the event log.  
The approach was evaluated using four real-life logs against two baselines for process variant analysis. The results show that at trace level, our approach reveals significant statistical discrepancies between the two variants, whereas the baselines used in our evaluation are unable to detect these differences. Furthermore, the presented approach performs within reasonable execution times, despite the more involving computations.

We foresee the applicability of the devised encoding technique based on discrete wavelet transformation to a range of process mining problems. These range from predictive process monitoring through to trace clustering, outlier detection, and process drift identification and characterization. 


\smallskip\noindent\textbf{Reproducibility} The source code required to reproduce the reported experiments can be found at \url{https://github.com/farbodtaymouri/RuleDeviance}.

\smallskip\noindent\textbf{Acknowledgments} This research is partly funded by the Australian Research Council (DP180102839) and Spanish funds MINECO and FEDER (TIN2017-86727-C2-1-R).

\bibliographystyle{elsarticle-num}
\bibliography{mybibfile}

\begin{thebibliography}{10}
\expandafter\ifx\csname url\endcsname\relax
  \def\url#1{\texttt{#1}}\fi
\expandafter\ifx\csname urlprefix\endcsname\relax\def\urlprefix{URL }\fi
\expandafter\ifx\csname href\endcsname\relax
  \def\href#1#2{#2} \def\path#1{#1}\fi

\bibitem{Suriadi2014}
S.~Suriadi, R.~S. Mans, M.~T. Wynn, A.~Partington, J.~Karnon, Measuring patient
  flow variations: A cross-organisational process mining approach, in: Asia
  Pacific BPM, Springer, 2014.

\bibitem{Swinnen2012}
J.~Swinnen, B.~Depaire, M.~J. Jans, K.~Vanhoof, A process deviation analysis --
  a case study, in: BPM Workshops, Springer, 2012, pp. 87--98.

\bibitem{Poelmans2010}
J.~Poelmans, G.~Dedene, G.~Verheyden, H.~Van~der Mussele, S.~Viaene, E.~Peters,
  Combining business process and data discovery techniques for analyzing and
  improving integrated care pathways, in: ICDM, Springer, 2010, pp. 505--517.

\bibitem{Aalst2016}
W.~M. P. v.~d. Aalst, {Process Mining: Data Science in Action}, 2nd Edition,
  Springer, 2016.

\bibitem{bolt2018}
A.~Bolt, M.~de~Leoni, W.~M. van~der Aalst, Process variant comparison: Using
  event logs to detect differences in behavior and business rules, Information
  Systems 74 (2018) 53--66.

\bibitem{Nguyen2018-Multiperspectquteprints117962}
H.~H. Nguyen, M.~Dumas, M.~L. Rosa, A.~H. ter Hofstede, Multi-perspective
  comparison of business process variants based on event logs (extended paper)
  (April 2018).

\bibitem{taymouri2019business}
F.~Taymouri, M.~L. Rosa, M.~Dumas, F.~M. Maggi, Business process variant
  analysis: Survey and classification (2019).
\newblock \href {http://arxiv.org/abs/1911.07582} {\path{arXiv:1911.07582}}.

\bibitem{Beest2015}
N.~R. T.~P. van Beest, M.~Dumas, L.~Garc{\'i}a-Ba{\~{n}}uelos, M.~La~Rosa, Log
  delta analysis: Interpretable differencing of business process event logs,
  in: BPM, Springer, 2015.

\bibitem{Cordes2015}
C.~Cordes, T.~Vogelgesang, H.-J. Appelrath, A generic approach for calculating
  and visualizing differences between process models in multidimensional
  process mining, in: BPM Workshops, Springer, 2015, pp. 383--394.

\bibitem{Ballambettu2017}
N.~P. Ballambettu, M.~A. Suresh, R.~P. J.~C. Bose, Analyzing process variants
  to understand differences in key performance indices, in: AISE, Springer,
  2017, pp. 298--313.

\bibitem{Kriglstein13}
S.~Kriglstein, G.~Wallner, S.~Rinderle-Ma, A visualization approach for
  difference analysis of process models and instance traffic, in: Business
  Process Management, Springer Berlin Heidelberg, 2013.

\bibitem{Pini2015}
A.~Pini, R.~Brown, M.~T. Wynn, Process visualization techniques for
  multi-perspective process comparisons, Springer, 2015, pp. 183--197.

\bibitem{WYNN201793}
M.~Wynn, E.~Poppe, J.~Xu, A.~ter Hofstede, R.~Brown, A.~Pini, W.~van~der Aalst,
  Processprofiler3d: A visualisation framework for log-based process
  performance comparison, DSS 100.

\bibitem{LOW2017106}
W.~Low, W.~van~der Aalst, A.~ter Hofstede, M.~Wynn, J.~D. Weerdt, Change
  visualisation: Analysing the resource and timing differences between two
  event logs, Inf. Systems 65.

\bibitem{Bolt2016}
A.~Bolt, M.~de~Leoni, W.~M.~P. van~der Aalst, A visual approach to spot
  statistically-significant differences in event logs based on process metrics,
  in: AISE, Springer, 2016.

\bibitem{strang16}
G.~Strang, Introduction to Linear Algebra, 5th Edition, Wellesley-Cambridge
  Press, 2016.

\bibitem{HaarWaveletApplication}
S.~{Santoso}, E.~J. {Powers}, W.~M. {Grady}, Power quality disturbance data
  compression using wavelet transform methods, IEEE Transactions on Power
  Delivery 12~(3) (1997) 1250--1257.

\bibitem{WaveletRecurrent}
K.~{Rao}, N.~{Ahmed}, Orthogonal transforms for digital signal processing, in:
  IEEE ICASSP '76, Vol.~1, 1976, pp. 136--140.

\bibitem{GallieLinearAlgebra2019book}
J.~Gallier, J.~Quaintance, Linear Algebra and Optimization with Applications to
  Machine Learning: Volume I, 2019.

\bibitem{Shumway:2303176}
R.~H. Shumway, D.~S. Stoffer, {Time series analysis and its applications: with
  R examples; 4th ed.}, Springer texts in statistics, Springer, Cham, 2017.

\bibitem{Aggarwal15}
C.~C. Aggarwal, Data Mining - The Textbook, Springer, 2015.

\bibitem{Bakshi1999MULTISCALEAA}
B.~R. Bakshi, Multiscale analysis and modeling using wavelets, 1999.

\bibitem{Yi2000}
B.{Yi}, N.~D. {Sidiropoulos}, T.~{Johnson}, H.~V. {Jagadish}, C.~{Faloutsos},
  A.~{Biliris}, Online data mining for co-evolving time sequences, in:
  Proceedings of 16th IEEE, ICDE, 2000.

\bibitem{Liu2007ComputationalMO}
H.~Liu, H.~Motoda, Computational methods of feature selection, 2007.

\bibitem{HORNIK1989359}
K.~Hornik, M.~Stinchcombe, H.~White, Multilayer feedforward networks are
  universal approximators, Neural Networks 2~(5) (1989) 359 -- 366.

\bibitem{Powers2011EvaluationFP}
D.~M.~W. Powers, Evaluation: from precision, recall and f-measure to roc,
  informedness, markedness and correlation, 2011.

\bibitem{wackerly08}
D.~D. Wackerly, W.~M. III, R.~L. Scheaffer, Mathematical Statistics with
  Applications, seventh edition Edition, Duxbury Advanced Series, 2008.

\bibitem{4dtu}
4{T}{U}: {C}entre for {R}esearch {D}ata, https://data.4tu.nl/repository (2019).

\end{thebibliography}

\end{document}